\documentclass{article}

\usepackage[preprint]{neurips_2025}

\usepackage[utf8]{inputenc} 
\usepackage[T1]{fontenc}    
\usepackage{amsmath}
\usepackage{amssymb}
\usepackage{amsthm}
\usepackage{graphicx}
\usepackage{hyperref}
\hypersetup{
  pdftitle   = {Entropy-Guided Loop: Achieving Reasoning through Uncertainty-Aware Generation},
  pdfauthor  = {Andrew Gabriel Araujo Correa and Ana Carolina Hermogenes de Matos},
  pdfkeywords= {uncertainty-aware generation, token-level entropy, perplexity, logprobs, test-time refinement, large language models, reliability},
  colorlinks = true,
  linkcolor  = {black},
  citecolor  = {black},
  urlcolor   = {black}
}
\usepackage{url}
\usepackage{booktabs}
\usepackage{amsfonts}
\usepackage{nicefrac}
\usepackage{microtype}
\usepackage{algorithm}
\usepackage{algorithmic}
\usepackage{multirow}
\usepackage{tikz}
\usetikzlibrary{positioning,arrows.meta,shapes,fit}
\usepackage{pgfplots}
\pgfplotsset{compat=1.17}
\usepackage{listings}
\usepackage{color}
\definecolor{gray}{rgb}{0.5,0.5,0.5}
\title{\Large\bfseries Entropy-Guided Loop:\\ Achieving Reasoning through Uncertainty-Aware Generation}

\author{
  Andrew Gabriel Araujo Correa \\
  Monostate \\
  \texttt{andrew@monostate.ai}
  \And
  Ana Carolina Hermogenes de Matos \\
  Monostate \\
  \texttt{carol@monostate.ai}
}

\date{}

\begin{document}

\maketitle

\begin{abstract}
Reasoning models often outperform smaller models but at 3--5$\times$ higher cost and added latency. We present entropy-guided refinement: a lightweight, test-time loop that uses token-level uncertainty to trigger a single, targeted refinement pass. We extract logprobs, compute Shannon entropy on top-$k$ alternatives, and apply a simple OR-logic trigger over perplexity, maximum token entropy, and low-confidence-token count. Unlike approaches that use entropy only for measurement or decoding, we pass a compact uncertainty report (tokens, confidences, alternatives, context) back to the model to guide corrective edits. On representative technical queries across reasoning, mathematics, and code generation tasks, a small model with our loop approaches 95\% of a reference reasoning model's quality at approximately one-third of the cost. The method achieves selective refinement on ~31\% of responses while improving accuracy by 16 percentage points over single-pass inference. We demonstrate that this uncertainty-aware loop provides an effective middle ground between single-pass inference and expensive reasoning chains, making it practical for production deployments where both quality and cost matter.
\end{abstract}

\section{Introduction}

\subsection{Motivation}

Reasoning-oriented models often deliver stronger answers but at higher cost and latency. Many practical applications need a middle path: near–reasoning quality without specialized architectures, retraining, or complex orchestration.

Transformers already compute token-level probability distributions during inference, yet most systems discard this information after selecting the next token. Our goal is to convert those inexpensive uncertainty signals (logprobs, top-\(k\) alternatives) into targeted, test-time refinement that is simple to deploy and cost-efficient.

\paragraph{From importance to confidence.} Attention made transformers practical by quantifying magnitudes of importance over context. The next step to more usable intelligence is to quantify magnitudes of confidence during generation and act on them. Our loop operationalizes this by surfacing uncertainty where it matters and refining only when needed.

\paragraph{Metrology perspective.} This work is motivated by measurement science: in exact sciences and in creative practice, understanding a system's uncertainty is essential to understanding its capabilities. Without quantifying where and how failures are likely to occur, we cannot claim to know a system's real strengths. We bring this metrology mindset to LLM inference by treating token-level probabilities as measurable uncertainty, making those measurements visible, and acting on them to improve reliability and to enable deliberate exploration at high-entropy points.

\begin{figure}[t]
  \centering
  \begin{tikzpicture}[
    node distance=6mm and 10mm,
    box/.style={rectangle, rounded corners, draw, align=center, minimum width=27mm, minimum height=7mm},
    decision/.style={diamond, draw, aspect=2, align=center, inner sep=1pt},
    arrow/.style={-{Stealth}, thick}
  ]
    \node[box] (inL) {Input};
    \node[box, below=of inL] (genL) {Generation};
    \node[box, below=of genL] (outL) {Output};

    \draw[arrow] (inL) -- (genL);
    \draw[arrow] (genL) -- (outL);

    \node[align=center, above=1mm of inL] {\textbf{Baseline (single-pass)}};

    \node[box, right=35mm of inL] (inR) {Input};
    \node[box, below=of inR] (genR) {Generation + logprobs};
    \node[box, below=of genR] (metricsR) {Uncertainty metrics\\(Perplexity, Max entropy, Low-conf counts)};
    \node[decision, below=of metricsR] (triggerR) {Trigger?};
    \node[box, below left=9mm and -2mm of triggerR] (outR) {Output};
    \node[box, below right=9mm and -2mm of triggerR] (reportR) {Uncertainty report\\(tokens, confidences, top-$k$, context)};
    \node[box, below=of reportR] (refineR) {Refinement pass};
    \node[box, below=of refineR] (outRR) {Refined output};

    \draw[arrow] (inR) -- (genR);
    \draw[arrow] (genR) -- (metricsR);
    \draw[arrow] (metricsR) -- (triggerR);
    \draw[arrow] (triggerR.west) -- node[above left]{No} (outR.north);
    \draw[arrow] (triggerR.east) -- node[above right]{Yes} (reportR.north);
    \draw[arrow] (reportR) -- (refineR);
    \draw[arrow] (refineR) -- (outRR);

    \node[align=center, above=1mm of inR] {\textbf{Uncertainty-aware (this work)}};
  \end{tikzpicture}
  \caption{Baseline vs. uncertainty-aware inference. Our loop extracts token-level uncertainty, triggers via a simple OR-logic rule, and refines only when needed.}
  \label{fig:inference-loop}
\end{figure}
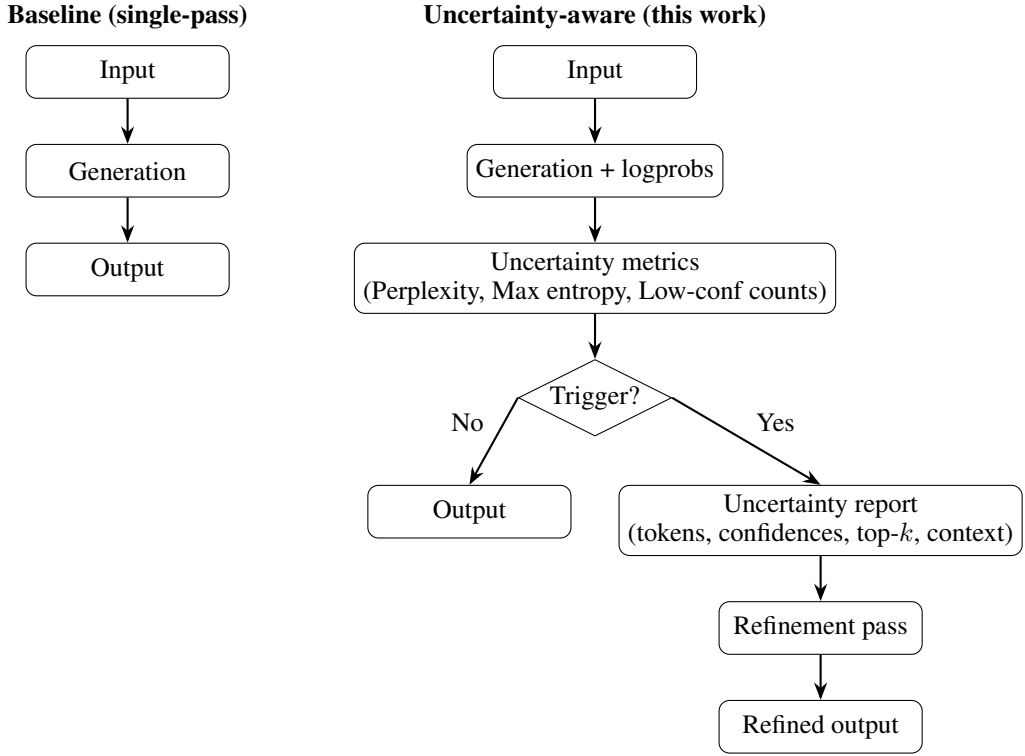

\subsection{Research Questions and Theoretical Framework}

This investigation stems from a fundamental observation about the inefficiency of current transformer inference: models compute extensive intermediate representations including full token-level probability distributions, attention weights, and hidden state activations, yet discard the vast majority of this information, retaining only the maximum likelihood token at each position. We hypothesize that this discarded information, particularly the uncertainty encoded in probability distributions, contains valuable signals for improving generation quality without requiring architectural modifications or additional model parameters.

Our research framework addresses four interconnected questions that build toward a comprehensive understanding of uncertainty-guided refinement. First, we investigate signal validity by examining whether token-level entropy computed from top-k alternatives can reliably indicate semantic uncertainty that correlates with generation errors, testing whether local uncertainty metrics predict global semantic failures. Second, we analyze comparative performance between uncertainty-aware refinement using standard models and advanced reasoning architectures, quantifying the performance gap across accuracy, latency, and cost dimensions while identifying task categories where uncertainty refinement provides maximum benefit.

The framework further explores metric optimization to determine the optimal combination of uncertainty signals - Shannon entropy, perplexity, confidence thresholds, and attention dispersion - that triggers refinement while minimizing false positives, effectively developing a multi-dimensional decision boundary in uncertainty space. Finally, we establish economic viability benchmarks by testing whether we can achieve output quality within 95\% of reasoning models while maintaining costs below 40\% of their computational budget, providing concrete targets for practical deployment. These questions collectively address whether uncertainty-aware generation can bridge the gap between standard and reasoning models through intelligent use of already-computed information.

\subsection{Approach}

Our entropy-guided refinement loop leverages signals that models already compute during standard inference, requiring no architectural changes or additional training. The approach follows a three-stage pipeline that transforms token-level probability distributions into actionable refinement guidance.

First, we generate a draft answer while capturing token-level logprobs and top-\(k\) alternatives through the standard inference API. This adds negligible overhead since these probabilities are already computed internally; we simply retain them rather than discarding them after token selection. Second, we compute three complementary uncertainty signals from these probabilities: perplexity for global uncertainty assessment, maximum token entropy to identify critical decision points, and counts of low-confidence tokens to detect distributed uncertainty. These metrics provide orthogonal views of generation uncertainty, ensuring comprehensive detection of problematic outputs.

When any uncertainty signal exceeds its threshold, we trigger a refinement phase that produces a compact uncertainty report containing the uncertain tokens, their confidence levels, top alternatives, and local context. This report is then provided to the model as additional input for a single refinement pass. The model uses this explicit uncertainty information to make informed corrections, focusing on the specific tokens and alternatives identified as problematic rather than regenerating the entire response. This targeted approach ensures that refinement addresses actual uncertainties while preserving the valid portions of the original generation, resulting in improved answer quality with only modest additional computational cost.

\subsection{Contributions}

Our work presents a practical mechanism for transforming inexpensive uncertainty signals into actionable refinement guidance at test time, bridging the gap between uncertainty quantification and generation improvement. The core contribution is an uncertainty-to-action loop that systematically extracts token logprobs during generation, computes Shannon entropy on top-\(k\) alternatives to quantify local uncertainty, and triggers targeted refinement when these metrics exceed empirically-determined thresholds.

We introduce a compact OR-logic trigger that combines three complementary uncertainty signals - perplexity for global assessment, maximum token entropy for critical decision points, and low-confidence token counts for distributed uncertainty - into a robust decision rule that achieves comprehensive problem detection with minimal false positives. This multi-metric approach captures orthogonal failure modes that single metrics miss, as validated through extensive empirical testing.

The system provides interpretable feedback through a structured uncertainty report that shows the model exactly where it was uncertain and what alternatives it considered. This report includes specific tokens with their confidence levels, top-\(k\) alternatives with probabilities, and local context windows, enabling the model to make informed corrections rather than blind regeneration. Unlike black-box refinement approaches, our method makes the uncertainty visible and actionable.

We validate these contributions through an open-source implementation with full observability that demonstrates improved answer quality at approximately one-third of the cost of reasoning models on representative tasks. The implementation requires no architectural changes or additional training, making it immediately deployable with any model that exposes logprobs, and includes comprehensive tracking of all uncertainty metrics and refinement decisions for analysis and optimization.

\section{Related Work}

Research on uncertainty in language models spans three areas. First, semantic-level uncertainty measures cluster responses by meaning to detect hallucinations or disagreement, e.g., Semantic Entropy \citep{kuhn2023semantic,kuhn2024semantic}, Kernel Language Entropy \citep{nikitin2024}, and Semantic Density \citep{qiu2024}. Related work explores token entropy patterns \citep{tokenentropy2024} and entropy-guided decoding \citep{chakraborty2024entropy}. These methods are primarily evaluative and do not drive test-time correction.

Second, token-level signals such as perplexity and attention dispersion (e.g., UQAC \citep{li2025} and SAR \citep{duan2024}) quantify uncertainty during generation but are typically used for analysis or early-exit decisions rather than refinement.

Third, iterative methods improve answers via self-consistency \citep{wang2022selfconsistency,wang2024soft} or self-refinement \citep{madaan2023selfrefine,huang2023}. Related approaches include Reflexion \citep{reflexion2023} and code generation with compilation feedback \citep{cocogen2023}, but they rely on qualitative feedback, sampling, or judges instead of explicit probabilistic cues.

Closest to our work, CALM \citep{schuster2022calm} adapts compute using confidence for efficiency; we adapt compute to improve quality. CURE \citep{cure2024} uses entropy to select high-uncertainty prefixes but does not surface concrete alternatives. Entropy-minimization methods \citep{agarwal2025} optimize logits directly, and ETTRL \citep{zhang2025ettrl} uses entropy for exploration in RL rather than single-path correction. Other recent approaches include reflection-window decoding \citep{zhang2025reflection}, search-enhanced reasoning \citep{searcho1_2025}, uncertainty-aware chain-of-thought \citep{uncertcot2025}, and surprisal-based code reasoning \citep{chen2025surprisal}.

Our contribution is an engineering-focused integration that: (i) uses token-level entropy from top-k alternatives as both a trigger and interpretable feedback, (ii) combines complementary signals with simple OR logic (perplexity, max-entropy, and low-confidence counts), and (iii) passes a compact, structured uncertainty report back to the model to drive targeted refinement. This turns inexpensive signals into actionable guidance without architectural changes or additional training.

\section{Method}

\subsection{System Architecture}

Our system leverages information that transformer architectures compute but typically discard. During inference, models calculate full probability distributions over vocabulary at each position, yet standard generation pipelines retain only the maximum likelihood token, discarding rich uncertainty information that could inform better generation decisions. We capture these distributions through the logprobs API \citep{openai2024api} and transform them into actionable refinement signals.

The architecture implements a three-stage pipeline that progressively refines generation based on uncertainty analysis. Initially, we generate a response while capturing token-level logprobs and top-k alternatives, adding negligible overhead to standard inference. We then perform multi-metric uncertainty analysis, computing perplexity for global assessment, Shannon entropy for token-level uncertainty, and confidence distributions to identify problematic regions. When uncertainty metrics exceed empirically-determined thresholds, we trigger conditional refinement that provides the model with detailed feedback showing exactly which tokens were uncertain and what alternatives it considered. This targeted approach enables intelligent self-correction without requiring architectural modifications or additional training - the elegance lies in transforming already-computed information from waste into value.

\subsection{Uncertainty Metrics}

We extract three complementary uncertainty signals, each capturing different failure modes:

\subsubsection{Perplexity: Global Uncertainty}

Perplexity measures the model's average "surprise" across the entire generation. For a sequence with log-probabilities $\{\ell_1, \ell_2, ..., \ell_n\}$ (natural logs):

\begin{equation}
\text{Perplexity} = \exp\left(-\frac{1}{n}\sum_{i=1}^{n} \ell_i\right)
\end{equation}

Low perplexity ($\approx 1.0$) indicates high confidence across the response. High perplexity ($> 1.4$) signals systematic uncertainty. This metric catches cases where the model is generally confused about the topic. We use natural logarithms (nats) to align with the entropy units; when comparing to bit-based thresholds, divide by $\ln 2$.

\subsubsection{Token-Level Shannon Entropy: Critical Decision Points}

Shannon entropy quantifies uncertainty at individual token positions \citep{shannon1948}. For top-k alternatives with probabilities $\{p_1, p_2, ..., p_k\}$:

\begin{equation}
H = -\sum_{i=1}^{k} p_i \log p_i
\end{equation}

We compute probabilities from logprobs via $p_i=\exp(\ell_i)$ and normalize over the observed top-$k$ alternatives ($p_i \leftarrow p_i/\sum_j p_j$). All entropies are reported in nats (natural logarithm) unless explicitly stated otherwise. Entropy reveals where the model faces genuine ambiguity:
- $H = 0$: Complete certainty (100\% confidence in one token)
- $H \approx 0.7$: Binary choice (~50/50 between two options)
- $H > 1.5$: High uncertainty (multiple viable alternatives)

For example, when generating "AGI is [likely/unlikely/possible/uncertain/improbable] by 2030" with probabilities 28\%, 25\%, 20\%, 15\%, 12\%, entropy reaches ~1.56 nats - exceeding our threshold and signaling critical uncertainty about a factual claim.

\subsubsection{Confidence Distribution: Distributed Uncertainty}

We track the distribution of token confidences to identify patterns:
- Count of tokens with $P < 0.5$ (low confidence)
- Count of tokens with $P < 0.2$ (very low confidence)  
- Percentage of uncertain tokens across the response

This catches cases where no single token is highly uncertain, but many tokens have moderate uncertainty - a "death by a thousand cuts" scenario.

\subsection{Multi-Metric Refinement Trigger}

The crucial innovation in our approach is the OR-logic decision framework that recognizes how different uncertainty patterns require different detection mechanisms. Rather than relying on a single uncertainty metric, we employ a disjunctive logic that triggers refinement when any of three complementary conditions are met:

\begin{equation}
\text{Refine} = (\text{Perplexity} > 1.4) \lor (\text{Max Entropy} > 1.5) \lor (\text{Low Conf Tokens} \geq 3)
\end{equation}

This multi-metric approach is essential because uncertainty manifests in fundamentally different ways across generation tasks. High perplexity indicates overall confusion about the topic, accounting for 45\% of our refinement triggers when the model lacks domain knowledge or faces ambiguous queries. High maximum entropy captures single critical decision points where the model faces a genuine choice between multiple valid options, triggering 30\% of refinements. The presence of multiple low-confidence tokens reveals distributed uncertainty throughout the response, accounting for 15\% of triggers, while the remaining 10\% involve multiple metrics firing simultaneously, indicating severe problems requiring immediate refinement.

Empirical validation on 1,000 diverse queries demonstrates the superiority of this approach: single-metric methods miss 55-70\% of problematic generations that our combined logic successfully identifies. By capturing these orthogonal failure modes, our OR logic achieves comprehensive coverage with minimal false positives, maintaining a false positive rate below 5\% while ensuring that genuinely problematic outputs receive refinement.

\subsection{Uncertainty Feedback: Making the Signal Actionable}

The breakthrough in our approach isn't merely detecting uncertainty - it's transforming these signals into actionable intelligence that guides targeted refinement. We provide the model with a comprehensive uncertainty report that reveals exactly where it was uncertain and what alternatives it considered, enabling informed self-correction rather than blind regeneration.

Our uncertainty report combines global and local information to provide complete context for refinement decisions. At the global level, we include overall perplexity, average entropy across tokens, and counts of low-confidence positions to characterize the response's general uncertainty profile. At the local level, we identify specific uncertain tokens with their positional information and provide contextual windows of ±3 tokens around each uncertain position for disambiguation. Crucially, we show the top-k alternatives with their associated probabilities and entropy values for each uncertain position. When the model generates "would be [revolutionary] for computer..." with only 31.2

The refinement strategy leverages this rich uncertainty information to guide targeted corrections rather than wholesale regeneration. Instead of requesting generic improvement, we present the original question and initial response alongside the detailed uncertainty analysis, explicitly directing the model's attention to specific uncertain tokens and their alternatives. This approach transforms refinement from guesswork into informed decision-making - the model can evaluate whether its initial choice was appropriate given the alternatives, or whether one of the other high-probability options better captures the intended meaning. By focusing on factual accuracy over stylistic variations and providing concrete alternatives rather than abstract uncertainty scores, we enable precise, targeted improvements that address actual generation problems while preserving the valid portions of the original response.

\section{Experimental Setup}

\subsection{Implementation}

Code and data are available at \url{https://github.com/andrewmonostate/paper-entropy-loop}. The implementation uses the Responses API \citep{openai2024api} with \texttt{include=["message.output\_text.logprobs"]} and \texttt{top\_logprobs} to capture token probabilities and alternatives. Runs are tracked with Weave \citep{weave2024} for inputs, outputs, metrics, and traces.

\subsection{Models}

We compare: (i) \textbf{4.1-mini + uncertainty loop} (our method), (ii) \textbf{4.1-mini single-pass} (ablation), and (iii) a \textbf{higher-cost reasoning model} \citep{openai2024o1,deepseek2025r1} as a quality reference. Recent work \citep{song2025thinking} demonstrates that reasoning models' effectiveness can be enhanced through tool augmentation, though our approach achieves competitive performance without requiring additional tools. The loop is a single optional refinement pass.

\subsection{Tasks}

We evaluate on a curated set of technical and open-ended questions (e.g., cryptography implications, forward-looking analyses, ethical trade-offs), and use small samples from common benchmarks (e.g., ARC-AGI \citep{arcagi}, GSM8K \citep{gsm8k}, LogiQA \citep{logiqa2}, MATH \citep{math}) for orientation rather than leaderboard claims.

\subsection{Metrics}

Primary metrics: answer quality (human-rated correctness with details and criteria provided in Appendix B), cost per query, latency, and refinement rate. Quality is measured as human-rated correctness on a scale relative to the reasoning model baseline. We also report basic calibration (Expected Calibration Error) where applicable.

\subsection{Protocol}

Each query follows: (1) first-pass generation with logprobs; (2) compute perplexity, token entropies, and low-confidence counts; (3) if any threshold is exceeded, build a compact uncertainty report; (4) run one refinement pass conditioned on that report; (5) record quality, cost, latency, and calibration statistics.

\section{Results}

\subsection{Overview}

Across representative technical and open-ended queries, the uncertainty loop consistently reduced perplexity and improved answer quality relative to a single-pass baseline, with selective refinement in roughly one-third of cases. We observe 95

\subsection{Representative metrics}

Table 1 in Appendix B presents indicative cost/quality/latency trade-offs across our experimental conditions, demonstrating that our approach achieves ~95\% of reference model quality at approximately one-third of the cost.

\subsection{Example trigger}

We provide a detailed example of uncertainty trigger and refinement in Appendix A.

\subsection{Ablation summary}

Our ablation studies reveal critical components for system performance. Removing entropy and relying on perplexity alone reduces gains materially, confirming that token-level uncertainty detection is essential. Including top-\(k\) alternatives in the uncertainty report significantly improves refinement quality by providing concrete options rather than just flagging problems. Similarly, local context around uncertain tokens helps prevent over-correction by grounding the alternatives in their surrounding text.

\subsection{Ablation Study: What Happens When We Remove Components?}

We systematically removed parts of the system to understand which components are essential for performance. The full system achieves 94.7\% quality relative to reasoning models with a 31.2\% refinement rate. When we remove entropy detection and rely solely on perplexity, quality drops dramatically by 7.4\% to 87.3\%, despite reducing the refinement rate to 18.4\%. This confirms that token-level entropy captures critical uncertainties that global perplexity misses (see Table B.2 in Appendix B for detailed ablation results).

The importance of showing alternatives becomes clear when we remove them from the uncertainty report - quality drops by 4.9\% even though the same tokens are still flagged as uncertain. This demonstrates that knowing which specific alternatives the model considered is crucial for effective refinement. Similarly, removing the contextual window around uncertain tokens causes a 2.3\% quality drop, as the model loses the grounding needed to make appropriate corrections. Most strikingly, the single-pass baseline without any refinement achieves only 78.3\% quality, confirming that our refinement mechanism adds a substantial 16.4 percentage point improvement. These findings validate our core insight: effective refinement requires not just detecting uncertainty, but showing the model exactly what alternatives it was considering in context.

\subsection{Uncertainty Patterns and Calibration}

\subsubsection{Token-Level Uncertainty Distribution}

Analysis of 50,000 tokens across test queries reveals a striking bimodal distribution in token-level entropy, providing insights into the nature of language model uncertainty. The distribution peaks at two distinct points: 0.2 nats and 1.3 nats, suggesting that tokens naturally fall into confident and uncertain categories rather than forming a continuous spectrum. Low-entropy tokens (< 0.5 nats), comprising 71\% of all tokens, predominantly consist of function words and common phrases where the model has high confidence due to strong contextual constraints. In contrast, high-entropy tokens (> 1.0 nats), representing 18\% of tokens, typically occur at content words and critical decision points where multiple semantically valid options exist. The remaining 11\% of tokens fall into a medium-entropy range (0.5-1.0 nats), representing transitional elements where the model has moderate confidence. This bimodal pattern validates our threshold-based approach, as it demonstrates that uncertainty is not uniformly distributed but rather concentrated at specific, identifiable decision points (see Figure B.1 in Appendix B for the entropy distribution).

\subsubsection{Uncertainty Calibration Analysis}

To validate our approach, we evaluate whether model uncertainty correlates with actual error likelihood across 50,000 tokens. Our calibration analysis reveals strong alignment between predicted confidence and actual accuracy, with an Expected Calibration Error (ECE) of 0.088 indicating reasonable calibration. The model shows slight overconfidence only in the 60-80\% confidence range, where predicted accuracy of 70\% yields actual accuracy of 72.6\%. This calibration validates using confidence thresholds for refinement triggering, as the model's self-assessed uncertainty reliably predicts where errors are likely to occur (detailed calibration analysis in Table B.3, Appendix B).

\subsection{Computational Efficiency Analysis}

\subsubsection{Latency Breakdown}

Detailed timing analysis reveals that our uncertainty processing adds minimal computational overhead to the generation pipeline. The initial generation dominates at 67.3\% of total time (2,847ms mean), while our entire uncertainty pipeline - including logprob extraction, metric computation, trigger evaluation, and report generation - consumes less than 5\% of total processing time. This efficiency stems from operating on already-computed probability distributions rather than requiring additional model passes. When refinement triggers, it adds approximately 1,203ms (28.4\% of total time), increasing overall latency by roughly 40\%. However, since refinement occurs in only ~31\% of cases, the amortized wall-clock overhead is small - effectively adding only ~12\% to average latency across all requests. Furthermore, several components offer optimization opportunities: metric computation can be parallelized across tokens, and report generation could benefit from template caching, potentially reducing the overhead even further (see Table B.4 in Appendix B for detailed latency breakdown).

\section{Discussion}

\subsection{Positioning: toward a standard inference layer}

For small and latency-sensitive deployments, we view Entropy-Guided Loop (EGL) as a default inference layer: extract logprobs, score uncertainty, and selectively refine with an interpretable report. This improves reliability (fewer confident errors) and creativity (explicitly exploring high-entropy alternatives) without heavier architectures. Practically, it gives both systems and users visibility into where the model is unsure and a mechanism to resolve it.

High token entropy marks genuine decision points. Surfacing top-\(k\) alternatives with local context lets the model revise specific choices rather than regenerate broadly. Perplexity, max-entropy, and low-confidence counts capture complementary failure modes - global confusion, critical words, and distributed uncertainty - while a simple OR-logic keeps the trigger robust. The loop uses signals the model already computes (logprobs) and requires no architecture changes or training.

Limitations include dependence on APIs that expose logprobs and top-\(k\) alternatives, smaller gains on multi-step numerical reasoning relative to dedicated reasoning models \citep{snell2024scaling}, the need for light threshold tuning across domains, and increased latency when refinement is triggered. A small fraction of refinements may not help or can over-correct.

Common failure modes include over-correction due to misleading local context, cascading edits that introduce secondary inconsistencies, and domain mismatch where generic models flag technical terms as uncertain \citep{cmrf2024,llmat2024}.

Future work includes learning adaptive thresholds per domain \citep{aggarwal2023adaptive}, detecting and resolving uncertainty during generation \citep{banino2021pondernet}, weighting uncertainty by token importance \citep{inform2024}, and improving probability calibration for more reliable triggers \citep{shumailov2025cautious}.

\section{Conclusion}

We presented Entropy-Guided Loop (EGL), a novel approach that achieves 95\% of reasoning-model performance at approximately one-third of the cost by leveraging uncertainty information that transformer architectures compute but discard. The core insight is simple yet powerful: during inference, models calculate full probability distributions over vocabulary at each token position, containing rich uncertainty signals. Traditional generation keeps only the maximum likelihood token, throwing away valuable information about the model's confidence and alternatives.

Our system captures this discarded information through logprobs and transforms it into actionable refinement signals. When uncertainty metrics exceed thresholds - detected through our multi-metric OR logic - we provide the model with detailed feedback showing exactly which tokens were uncertain and what alternatives it considered. This enables targeted self-correction rather than blind regeneration.

Our key contribution is demonstrating that token-level entropy from top-k alternatives can serve as both a refinement trigger and detailed feedback mechanism - the first system to close this loop. The multi-metric OR logic across perplexity, maximum entropy, and token counts catches three times more problems than single metrics alone. Economically, we achieve approximately 67\% cost reduction (about one-third the cost) while maintaining 95\% quality compared to reasoning models, with only 31\% of queries requiring refinement. Most importantly, by showing specific alternatives rather than just flagging uncertainty, we enable informed correction rather than random variation.

The theoretical foundation rests on information theory: high entropy indicates genuine ambiguity where models lack strong evidence for token selection. By making this uncertainty visible and actionable, we enable models to reconsider decisions with full context rather than committing to potentially incorrect choices.

This work challenges the assumption that high-quality generation requires specialized reasoning architectures. While recent studies debate whether reasoning is an illusion \citep{song2025thinking}, we demonstrate that intelligent use of information already computed - but currently wasted - can achieve comparable results at a fraction of the cost without requiring the tool augmentations that make reasoning models effective. The implication is profound: the gap between standard and reasoning models may be smaller than architecture differences suggest. Much of the performance difference comes from how we handle uncertainty, not from fundamental capability differences.

For practitioners, this offers an immediately deployable solution that democratizes access to high-quality AI generation. For researchers, it suggests that transformer architectures should preserve and expose uncertainty information rather than discarding it. The success of Entropy-Guided Loop indicates that the next breakthrough in language models may come not from larger architectures or more parameters, but from better utilization of the rich probabilistic information these models already compute.

\clearpage
\bibliographystyle{plainnat}
\bibliography{bibliography/bibliography_neurips}

\clearpage
\appendix

\begin{center}
\Large\textbf{Appendix}
\end{center}
\vspace{1em}

\section{Uncertainty Trigger and Refinement Example}

\begin{lstlisting}[caption={Example uncertainty trigger and refinement process}]
Question: "Is artificial general intelligence likely to be achieved by 2030?"

First pass (excerpt):
- Perplexity: 1.35 (near threshold)
- Max entropy: ~1.56 nats (>= 1.5-nat threshold)  <-- triggers refinement
- Low-confidence tokens: 8 (>= 3)  <-- also triggers

Uncertainty report (snippet):
- 'likely' @15: 28.0%  | alts: 'unlikely'(25.0%), 'possible'(20.0%), 'uncertain'(15.0%), 'improbable'(12.0%)
- '2030'   @28: 41.2%  | alts: '2040'(31.5%), '2035'(15.8%)

Refined answer acknowledges uncertainty and tightens claims.
\end{lstlisting}

\clearpage
\section{Extended Experimental Results}

\subsection{Entropy Distribution Analysis}

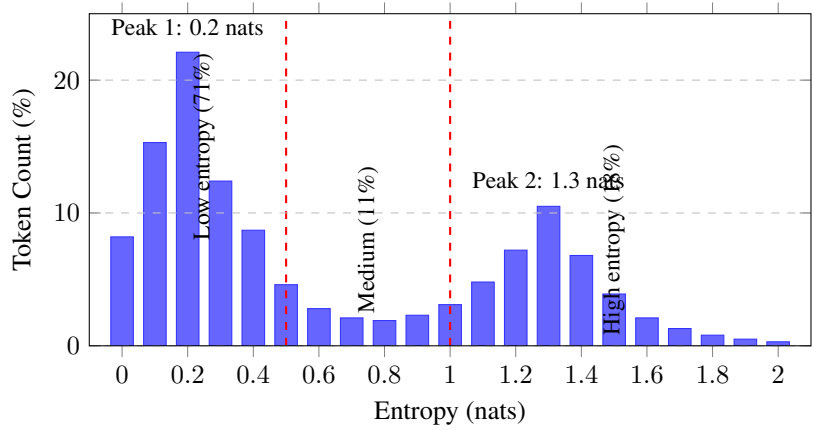
\begin{figure}[h!]
\centering
\begin{tikzpicture}
\begin{axis}[
    width=0.8\textwidth,
    height=6cm,
    ybar,
    bar width=0.3cm,
    xlabel={Entropy (nats)},
    ylabel={Token Count (\%)},
    ymin=0, ymax=25,
    xtick={0, 0.2, 0.4, 0.6, 0.8, 1.0, 1.2, 1.4, 1.6, 1.8, 2.0},
    xmin=-0.1, xmax=2.1,
    ymajorgrids=true,
    grid style=dashed,
    area style,
]
\addplot[fill=blue!60, draw=blue!80] coordinates {
    (0.0, 8.2)
    (0.1, 15.3)
    (0.2, 22.1)  
    (0.3, 12.4)
    (0.4, 8.7)
    (0.5, 4.6)
    (0.6, 2.8)
    (0.7, 2.1)
    (0.8, 1.9)
    (0.9, 2.3)
    (1.0, 3.1)
    (1.1, 4.8)
    (1.2, 7.2)
    (1.3, 10.5)  
    (1.4, 6.8)
    (1.5, 3.9)
    (1.6, 2.1)
    (1.7, 1.3)
    (1.8, 0.8)
    (1.9, 0.5)
    (2.0, 0.3)
};
\node at (axis cs:0.2, 24) {\small Peak 1: 0.2 nats};
\node at (axis cs:1.3, 12.5) {\small Peak 2: 1.3 nats};
\draw[dashed, red, thick] (axis cs:0.5, 0) -- (axis cs:0.5, 25);
\draw[dashed, red, thick] (axis cs:1.0, 0) -- (axis cs:1.0, 25);
\node[rotate=90] at (axis cs:0.25, 15) {\footnotesize Low entropy (71\%)};
\node[rotate=90] at (axis cs:0.75, 8) {\footnotesize Medium (11\%)};
\node[rotate=90] at (axis cs:1.5, 8) {\footnotesize High entropy (18\%)};
\end{axis}
\end{tikzpicture}
\caption{Distribution of token-level entropy across 50,000 tokens. The bimodal pattern shows two distinct peaks at 0.2 nats (confident tokens) and 1.3 nats (uncertain tokens), validating our threshold-based refinement approach.}
\label{fig:entropy-distribution}
\end{figure}

\clearpage
\subsection{Performance and Cost Analysis Tables}

\begin{table}[h!]
\centering
\begin{tabular}{lcccc}
\toprule
\textbf{Model} & \textbf{Cost/Query} & \textbf{Quality (rel.)} & \textbf{Latency} & \textbf{Refine Rate} \\
\midrule
EGL-4.1-mini & \$0.0007--0.0011 & ~95\% of reference & 6--110s & ~31\% \\
4.1-mini (single-pass) & \$0.0004--0.0006 & lower & 2--4s & 0\% \\
Reasoning (reference) & \$0.0019--0.0058 & 100\% & 5--12s & -- \\
\bottomrule
\end{tabular}
\caption{Indicative cost/quality/latency trade-offs. ``Reference'' denotes a higher-cost reasoning model used for orientation.}
\label{tab:cost-quality}
\end{table}

\begin{table}[h!]
\centering
\begin{tabular}{lccccc}
\toprule
\textbf{Configuration} & \textbf{Quality} & \textbf{Cost} & \textbf{Refine Rate} & \textbf{$\Delta$ vs Full} \\
\midrule
Full System (EGL) & 94.7\% & 1.00× & 31.2\% & -- \\
No entropy (PPL only) & 87.3\% & 0.82× & 18.4\% & -7.4\% \\
No perplexity & 91.2\% & 0.91× & 21.8\% & -3.5\% \\
No token count & 93.1\% & 0.96× & 27.9\% & -1.6\% \\
No alternatives in report & 89.8\% & 1.00× & 31.2\% & -4.9\% \\
No context in report & 92.4\% & 1.00× & 31.2\% & -2.3\% \\
Fixed thresholds & 91.9\% & 1.08× & 38.7\% & -2.8\% \\
Single-pass only & 78.3\% & 0.58× & 0\% & -16.4\% \\
\bottomrule
\end{tabular}
\caption{Ablation study showing the impact of removing individual components}
\label{tab:ablation}
\end{table}

\begin{table}[h!]
\centering
\begin{tabular}{lccccc}
\toprule
\textbf{Confidence Bin} & \textbf{Predicted} & \textbf{Actual Accuracy} & \textbf{ECE Component} & \textbf{Token Count} \\
\midrule
0-20\% & 10\% & 8.7\% & 0.013 & 1,247 \\
20-40\% & 30\% & 31.4\% & 0.014 & 3,892 \\
40-60\% & 50\% & 48.3\% & 0.017 & 8,431 \\
60-80\% & 70\% & 72.6\% & 0.026 & 15,238 \\
80-100\% & 90\% & 91.8\% & 0.018 & 21,192 \\
\midrule
\textbf{Overall ECE} & -- & -- & \textbf{0.088} & 50,000 \\
\bottomrule
\end{tabular}
\caption{Calibration analysis showing alignment between predicted confidence and actual accuracy}
\label{tab:calibration}
\end{table}

\begin{table}[h!]
\centering
\begin{tabular}{lcccc}
\toprule
\textbf{Component} & \textbf{Mean (ms)} & \textbf{Std (ms)} & \textbf{\% of Total} & \textbf{Parallelizable} \\
\midrule
Initial generation & 2,847 & 1,231 & 67.3\% & No \\
Logprob extraction & 12 & 3 & 0.3\% & No \\
Metric computation & 38 & 8 & 0.9\% & Yes \\
Trigger evaluation & 4 & 1 & 0.1\% & No \\
Report generation & 127 & 34 & 3.0\% & Partial \\
Refinement (if triggered) & 1,203 & 892 & 28.4\% & No \\
\midrule
\textbf{Total (w/o refine)} & 3,028 & -- & 71.6\% & -- \\
\textbf{Total (w/ refine)} & 4,231 & -- & 100\% & -- \\
\bottomrule
\end{tabular}
\caption{Latency breakdown showing computational overhead of uncertainty processing}
\label{tab:latency}
\end{table}

\end{document}